\documentclass[letterpaper,10pt,conference]{ieeeconf}
\IEEEoverridecommandlockouts

\usepackage[numbers,sort&compress]{natbib}
\usepackage{graphicx}
\usepackage{epsfig}
\usepackage{verbatim}
\usepackage{color}
\usepackage{soul}
\usepackage[font=footnotesize]{caption}
\usepackage{amssymb}
\usepackage{amsmath}
\usepackage{algorithm}
\usepackage{algpseudocode}
\usepackage[english]{babel}
\usepackage[T1]{fontenc}
\usepackage[utf8]{inputenc}
\usepackage{xspace}
\usepackage{todonotes}
\usepackage{epstopdf}
\usepackage{hyperref}
\usepackage{url}
\usepackage{physics}
\usepackage[nolist,nohyperlinks]{acronym}
\usepackage{tikz}
\usepackage{pgfplots}
\usepackage{subcaption}
\usepackage{siunitx}
\usepackage{makecell}

\usetikzlibrary{shapes,arrows,positioning}
\usetikzlibrary{calc}
\tikzstyle{block} = [draw, fill=blue!20, rectangle, 
    minimum height=3em, minimum width=6em]
\tikzstyle{sum} = [draw, fill=blue!20, circle, node distance=1cm]
\tikzstyle{input} = [coordinate]
\tikzstyle{output} = [coordinate]
\tikzstyle{pinstyle} = [pin edge={to-,thin,black}]

\title{3D Registration of Aerial and Ground Robots for Disaster Response: \\An Evaluation of Features, Descriptors, and Transformation Estimation}

\author{\authorblockN{
Abel Gawel\authorrefmark{1},
Renaud Dub\'e\authorrefmark{1},
Hartmut Surmann\authorrefmark{2},\\
Juan Nieto\authorrefmark{1}, 
Roland Siegwart\authorrefmark{1} and
Cesar Cadena\authorrefmark{1}
}
\authorblockA{\authorrefmark{1}Autonomous Systems Lab, ETH Zurich,
 \authorrefmark{2}Fraunhofer IAIS, University of Applied Sciences Gelsenkirchen}}

\begin{document}

\maketitle
\thispagestyle{empty}
\pagestyle{empty}

%

\maketitle

\begin{abstract}
Global registration of heterogeneous ground and aerial mapping data is a challenging task.
This is especially difficult in disaster response scenarios when we have no prior information on the environment and cannot assume the regular order of man-made environments or meaningful semantic cues.
In this work we extensively evaluate different approaches to globally register UGV generated \emph{3D} point-cloud data from LiDAR sensors with UAV generated point-cloud maps from vision sensors. 
The approaches are realizations of different selections for: a) local features: key-points or segments; b) descriptors: FPFH, SHOT, or ESF; and c) transformation estimations: RANSAC or FGR.
Additionally, we compare the results against standard approaches like applying ICP after a good prior transformation has been given.
The evaluation criteria include the distance which a UGV needs to travel to successfully localize, the registration error, and the computational cost.
%
In this context, we report our findings on effectively performing the task on two new Search and Rescue datasets.
Our results have the potential to help the community take informed decisions when registering point-cloud maps from ground robots to those from aerial robots.

\end{abstract}

\IEEEpeerreviewmaketitle

\section{Introduction}
\begin{figure}
\centering
\includegraphics[width = \columnwidth]{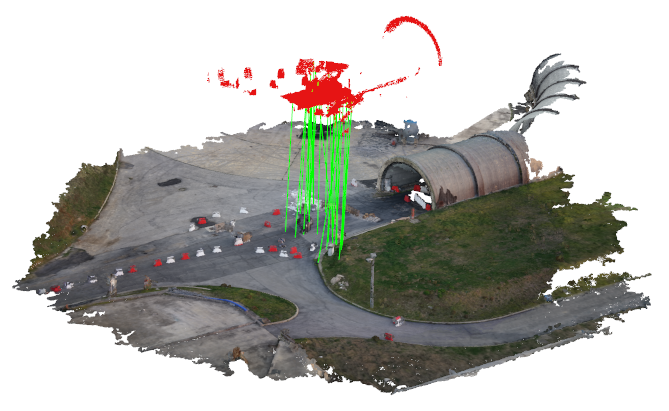}
\caption{Global localization of a \emph{3D} \ac{UGV} sub-map (red point-cloud) in a \emph{3D} \ac{UAV} reference map (coloured point-cloud). Green lines indicate the resulting matches associated to points in the two point-clouds. The data stems from the Montelibretti outdoor dataset and drawn from the complex experimental set-up.}
\label{fig:montelibretti}
\vspace{-5mm}
\end{figure}

Multi-robot applications with heterogeneous robotic teams are an increasing trend due to numerous advantages.
An \ac{UGV} can often carry high payloads and operate for extended periods of time, while an \ac{UAV} offers swift deployment and the opportunity to rapidly survey large areas.
This is especially beneficial in \ac{SaR} scenarios, see Fig.~\ref{fig:montelibretti} as an example.
Here, an initial overview can be made using \ac{UAV}s before deploying \ac{UGV}s for closer exploration in areas of interest.
However, when it comes to efficiently combining the strengths of such robotic teams, we face numerous challenges.
Additionally to the large difference in point of view, the sensor modalities used for mapping and localization are often drastically different for \ac{UAV}s and \ac{UGV}s.
While \ac{UAV}s typically use cameras as the prime sensor, \ac{UGV}s often rely on LiDAR.
This poses a major challenge in efficiently exploiting the \ac{UAV} data on a \ac{UGV} as registration between different sensor modalities is difficult to perform.
Furthermore, using advanced functionalities such as traversability analysis and path planning for \ac{UGV}s on \ac{UAV} generated maps requires tight alignment between the data of different modalities, active localization and suitable map representations.

One critical step in using maps across several robots is the identification of the alignment between their maps.
Several techniques are possible with increasing generality~\cite{saeedi2016multiple}.
Firstly it is possible to impose a common origin of different robots' maps, e.g., by using common starting locations as done by~\citet{michael2012collaborative}. 
Another option is to use global positioning sensors that allow for a good initial guess on the alignment of coordinate frames.
In the case that several robots operate concurrently, it is also possible to find an alignment by a relative localization of the robots against each other~\cite{Kim10icra}.
%
However, the most challenging task is to register maps without any prior information regarding their mutual alignment. 
Furthermore, for the \ac{SaR} application as in the \ac{TRADR} project, the scenarios are completely unpredictable which rules out the possibility of using supervised learning into the pipeline~\cite{zeng2017cvpr}.

While our previous work on online multi-robot SLAM for \emph{3D} LiDARs~\cite{Dube2017iros}\footnote{Within this paper, this system will be referred to as LaserSLAM.} demonstrates a reliable registration among point-cloud maps taken from multiple ground robots, there is still the issue of dealing with differences in modality and in point of view between \ac{UAV}s and \ac{UGV}s.

The above mentioned challenges motivate us to evaluate several techniques to globally localize a \ac{UGV} using its LiDAR sensor in a point-cloud map generated using the \ac{MVE}~\cite{fuhrmann2014mve} from images recorded by a \ac{UAV}.
The global registration (or localization) pipeline schematized in Fig.~\ref{fig:overview} consists of feature extraction, feature description and matching, and a 3D transformation estimation.
We provide an evaluation and an analysis of the implementation and performance of different choices for the modules in this registration pipeline.
These choices are:

\begin{itemize}
\item Local feature extraction: key-points or segments.
\item Feature descriptors: \ac{FPFH}, \ac{SHOT} or \ac{ESF}.
\item Transformation estimation: RANSAC based or \ac{FGR}.
\end{itemize}

The evaluation is conducted on two real world datasets of an indoor and an outdoor \ac{SaR} scenario.
This paper presents the following contributions:
\begin{itemize}
\item Extensive evaluation of global registration realizations for registering \ac{UGV} and \ac{UAV} point-clouds from LiDAR and camera data respectively.
\item Two new datasets for multi-modal SLAM in \ac{SaR} scenarios.
\end{itemize}

\section{Related work}
The field of \emph{2D} metrical map-merging based on overlapping map segments is well studied in literature \cite{birk2006merging, blanco2013robust, saeedi2011multiple}.
However, the task is increasingly difficult when moving to \emph{3D} environments \cite{saeedi2016multiple}, especially when dealing with heterogeneous robotic teams, where \emph{3D} data is generated from different sensors and with different noise characteristics \cite{Cadena16tro-SLAMfuture}.
\citet{michael2012collaborative} demonstrate a system for collaborative \ac{UAV}-\ac{UGV} mapping.
The authors propose a system where a \ac{UGV} equipped with a LiDAR sensor performs \emph{2.5D} mapping, using the flat ground assumption and consecutively merging scans using \ac{ICP}.
In dedicated locations a \ac{UAV} equipped with a 2D LiDAR is launched from the \ac{UGV} and maps the environment using a pose-graph SLAM algorithm.
Maps generated from the \ac{UAV} are then fused online with the \ac{UGV} map using \ac{ICP} initialized at the \ac{UAV} starting location. 

\citet{forster2013air} go a step further in fusing \ac{UAV}-\ac{UGV} map data from different sensors, i.e., RGB-D maps from the \ac{UGV} and dense monocular reconstruction from the \ac{UAV}.
The registration between the maps is performed using a \emph{2D} local height map fitting in \emph{x} and \emph{y} coordinates with an initial guess within a $3m$ search radius.
The orientation is a priori recovered from the magnetic north direction as measured by the \ac{IMU}s.
In a related setting \citet{Hinzmann2016} evaluate different variants of \ac{ICP} for registering dense \emph{3D} LiDAR point-clouds and sparse \emph{3D} vision point-clouds from \ac{SfM} recorded with different \ac{UAV}s into a common point-cloud map using an initial GPS prior for the map alignment.

Instead of using the generated \emph{3D} data for localizing between RGB and \emph{3D} LiDAR point-cloud data, \citet{wolcott2014visual} propose to generate \emph{2D} views from the LiDAR point-clouds based on the surface reflectivity.
However, this work focuses only on localization and it is demonstrated only on maps recorded from similar points of view.

In our previous work \cite{gawel2016structure} we presented a global registration scheme between sparse \emph{3D} LiDAR maps from \ac{UGV}s and vision keypoint maps from \ac{UAV}s, exploiting the rough geometric structure of the environment.
Here, registration is performed by clustering of geometric keypoint descriptors matches between map segments under the assumption of a known \emph{z}-direction as determined by an \ac{IMU}.

\citet{zeng2017cvpr} present geometric descriptor matching based on learning.
However, this approach is infeasible in unknown \ac{SaR} scenarios, as the descriptors do not generalize well to unknown environments.

\citet{dube2017icra} demonstrate better global localization performance in \emph{3D} LiDAR point-clouds by using segments as features instead of key-points.
This approach has been demonstrated with multiple \ac{UGV}s with the same robot-sensor set-up, but it is still to be studied how the approach performs under large changes in point of view.

Assuming good initialization of the global registration, \citet{zhou2016fast} perform a robust optimization.
The work claims faster and more robust performance than ICP.

In summary, the community addresses the problem of heterogeneous localization.
However, there is a research gap in globally localizing from one sensor modality to the other in full \emph{3D} without strong assumptions on view-point, terrain or initial guess.
\section{Aerial-Ground robot mapping system}
In this section, we present our SLAM system.
It extends our LaserSLAM system \cite{Dube2017iros} with the component of global map alignment and localization in point-cloud maps from different sources, as well as online extension of these maps. 
Fig.~\ref{fig:overview} illustrates the architecture of the proposed system.
While the major LaserSLAM system is running on the \ac{UGV}, it also allows to load, globally align, and use point-cloud maps from other sources. 
As example, maps generated via \ac{MVE} from \ac{UAV}s or point-cloud maps resulting from bundle adjustment on data collected by another LiDAR equipped robot can all be leveraged.
\pgfdeclarelayer{background}
\pgfdeclarelayer{foreground}
\pgfsetlayers{background,main,foreground}
\begin{figure*}
\centering
\begin{tikzpicture}[block/.style   ={rectangle, draw, text width=4.2em, text centered, rounded corners, minimum height=3em, minimum height=3em},
node distance=2.1cm, auto, >=latex']

	\node [block, align=center] (ugv) {\footnotesize UGV LiDAR\\ \footnotesize mapper};
	\node [block, align=center, right of=ugv] (ugv_map) {\footnotesize Local UGV\\ \footnotesize point-cloud};
	\node [block, align=center, below = 1cm of ugv_map] (keypoint) {\footnotesize Feature \\ \footnotesize extraction};
	\node [block, align=center, below = 1cm of keypoint] (uav_map) {\footnotesize Global UAV\\ \footnotesize point-cloud};
	\node [block, align=center, left of=uav_map] (uav) {\footnotesize UAV MVE\\ \footnotesize mapper};
	\node [block, align=center, right of=keypoint] (descriptors) {\footnotesize Descriptors};
	\node [block, align=center, right of=descriptors] (matching) {\footnotesize Descriptor \\ \footnotesize matching};
	\node [block, align=center, right of=matching] (consistency) {\footnotesize Transfor. \\ \footnotesize estimation};    
	\node [block, align=center, right=1cm of consistency] (icp) {\footnotesize ICP};
	\node [block, align=center, right of=icp] (map) {\footnotesize Fused point-\\ \footnotesize cloud map};
\node [align=center, right of=map] (output) {\footnotesize traversability\\ \footnotesize visualization\\ \footnotesize etc.};
	
\draw[->] (keypoint.east) +(0,-1em) coordinate (b1) -- (descriptors.west |- b1);
\draw[->] (keypoint.east) +(0,1em) coordinate (b1) -- (descriptors.west |- b1);
\draw[->] (descriptors.east) +(0,-1em) coordinate (b1) -- (matching.west |- b1);
\draw[->] (descriptors.east) +(0,1em) coordinate (b1) -- (matching.west |- b1);
\draw [->] (matching) -- node {} (consistency);	
\draw [->] (consistency) -- node {$\boldsymbol{T}_{UAV}^{UGV}$} (icp);
\draw [->] (icp) -- node {} (map);
\draw [->] (ugv) -- node {} (ugv_map);	
\draw [->] (uav) -- node {} (uav_map);	
\draw [->] (ugv_map) -- node {} (keypoint);
\draw [->] (uav_map) -- node {} (keypoint);
\draw [->] (ugv_map) -| node {} (icp);
\draw [->] (uav_map) -| node {} (icp);				
\draw [->] (map) -- node {} (output);

\begin{pgfonlayer}{background}
\path (keypoint.north west)+(-0.2,0.5) node (a) {};
\path (consistency.south -| consistency.east)+(+0.2,-0.6) node (b) {};
\path[rounded corners, draw=red, dashed]
    (a) rectangle (b);
    \path (matching.south) +(0,-0.4) node (glob) {Global registration};
\end{pgfonlayer}

\end{tikzpicture}
\caption{Mapping System overview. The inputs to the system are the local \ac{UGV} map and the global \ac{UAV} reference map. If a global registration is triggered, the key-points are computed on both maps. Consecutively, the system performs descriptor extraction and matching. The initial global transformation is then refined by a step of ICP between the global and the local clouds, resulting in a fused map that is used for further functionalities of the system, such as path planning.}
\label{fig:overview}
\end{figure*}
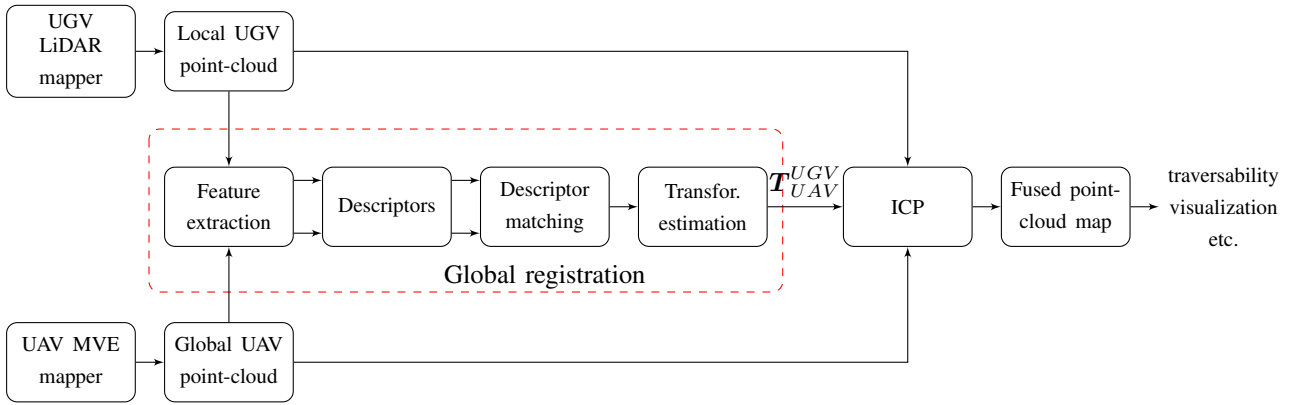
\subsection{Mapping algorithms}
\label{sec:mapping_algorithms}
In the proposed system, we use a dual map representation for the different tasks of the robots, i.e., point-cloud maps and \emph{OctoMaps} \cite{hornung2013octomap}.
While the individual robots maintain point-clouds and surface meshes, these are integrated in a global \emph{OctoMap} representation serving as the interface to other modules of the \ac{SaR} system, e.g., traversability analysis as shown in \cite{dube20163d}.
Another advantage of the unified \emph{OctoMap} representation is a persistent representation which also incorporates dynamic changes detection.

On the \ac{UAV}'s monocular image data we perform an \ac{SfM} and Multi-View-Stereo-based scene reconstruction using the \ac{MVE} \cite{fuhrmann2014mve}.
The \ac{MVE} produces a dense surface mesh of the scene by extensive matching and is therefore an offline method that is computed off-board the \ac{UAV}.
Although, efficient online mapping methods exist, we decide to use a method that produces high quality maps, that can be further used in the TRADR system, e.g., on the \ac{UGV} for path planning or for situation awareness of first responders.

On the \ac{UGV}s we use a variant of the LaserSLAM system which estimates in real-time the robot trajectory alongside with the \emph{3D} point-cloud map of the environment.
LaserSLAM is based on the \emph{iSAM2}~\cite{kaess2012isam2} pose-graph optimization approach and implements different types of sequential and place recognition constraints.
In this work, odometry constraints are obtained by fusing wheel encoders and IMU data using an extended Kalman filter while scan-matching constraints are obtained based on \ac{ICP} between successive scans.  

After the creation of the \ac{UAV} map, we facilitate a global registration scheme to localize the \ac{UGV} in the \ac{UAV} map as described in Sec.~\ref{sec:global_registration}.

For \ac{UGV}-only mapping, the LaserSLAM framework enables multiple robots to create consistent \emph{3D} point-cloud maps.
However, a different regime must be followed for generating and extending a consistent \emph{3D} map by fusing in the \ac{UAV} dense \emph{3D} maps.
Since the \ac{UAV} maps are the result of an offline batch optimization process, the maps are already loop closed and represent a consistent initial basis for the global map.
Furthermore, we treat the \ac{UAV} maps as static, i.e., the map is taken \emph{as is}.
LaserSLAM is therefore extended to include a mode that allows the robot to use a given base map.
This base map is then treated as the fully optimized map and extended with updates from the \ac{UGV} LiDAR.
It is important to note that the point-cloud map is only the internal representation for the robot to perform SLAM.
All map updates as well as dynamic changes are maintained in the \emph{OctoMap} representation that is derived from the point-cloud updates and serves as a unified representation for all processes using the mapping data.
\subsection{Map usage}
\begin{figure}
\centering
\includegraphics[width = 0.95\columnwidth]{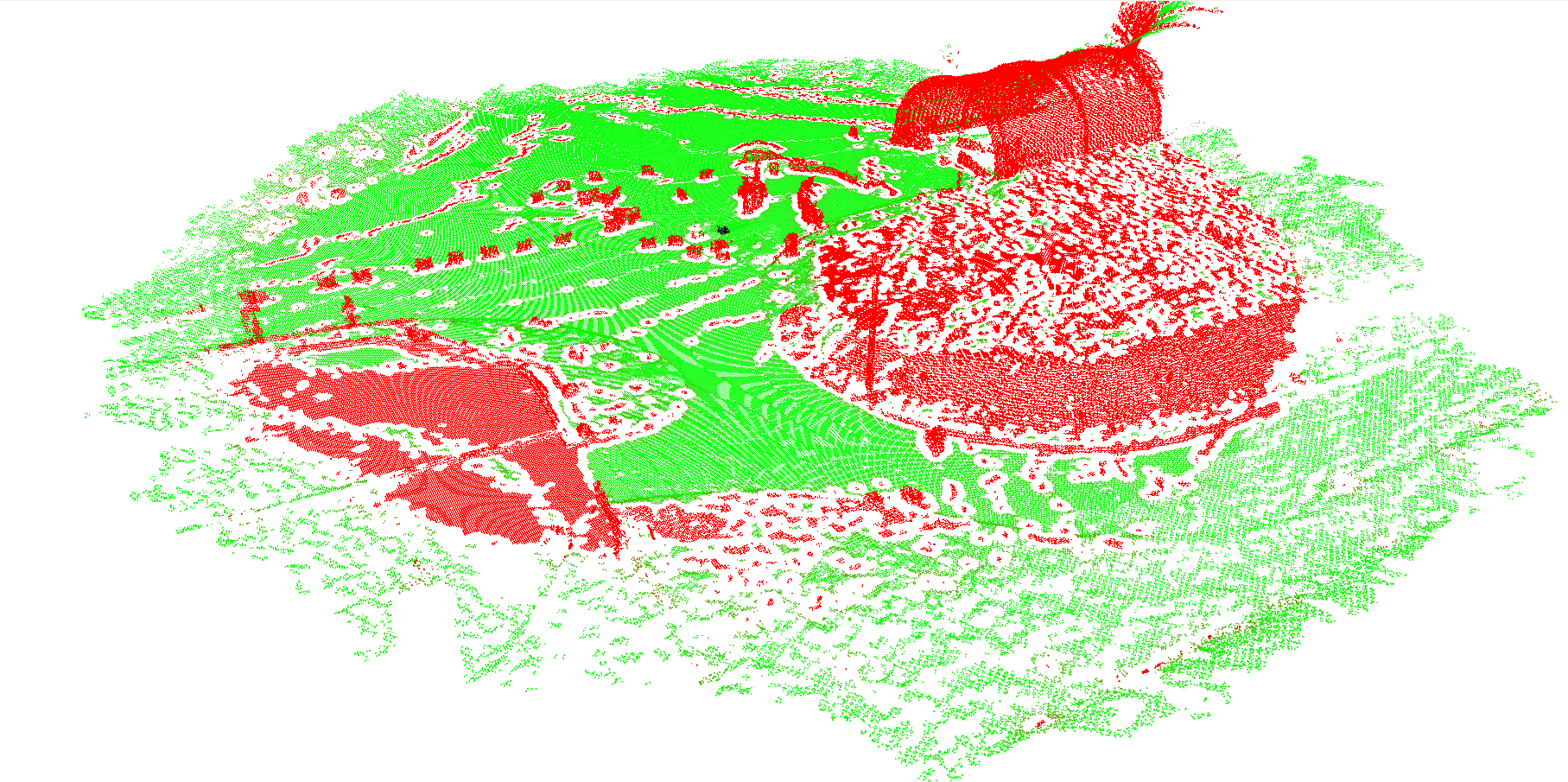}
\caption{Resulting \ac{UGV} traversability estimation on the outdoor Montelibretti dataset. Traversable areas are marked in green, while non traversable areas are indicated with red. The parametrization is identical to the parametrization for LiDAR traversability analysis.}
\label{fig:traversability}
\end{figure}

Thanks to the unified \emph{OctoMap} interface, the merged map data can directly be used on other modules of the TRADR system.
Notably, it can directly be used for traversability analysis and subsequent metrical path planning.
The system therewith enables the \ac{UGV}s to also use \ac{UAV} generated maps for path planning.
A \ac{UGV}-loaded \ac{UAV} map of one testing site (see Fig.~\ref{fig:montelibretti}) is depicted in Fig.~\ref{fig:traversability} indicating traversable areas in green and non-traversable areas in red.

The \emph{OctoMap} serves as interface for further modules of the system, such as novelty detection which is a separate contribution and out of the scope of this work.

\section{Global Registration}
\label{sec:global_registration}
This section describes the pipeline that we use to globally register the \ac{UGV} with respect to the \ac{UAV} point-cloud map.
The evaluation of different choices within this pipeline is the focus of our paper.
The global registration consist of four modules: feature extraction, description, matching and, estimation of the $\mathbb{SE}(3)$ transformation.

\subsection{Feature extraction}
This module defines which components in the point-cloud map are going to be used for the registration.
Key-point are samples from the full point-cloud that have some level of invariance to the point of view.
Here, we use the \ac{ISS} detector~\cite{zhong2009intrinsic}.
The next option, is to add more information by clustering the point-cloud, resulting in segments.
These segments are taken as the local features with the potential of being more descriptive than just 3D points~\cite{dube2017icra}.
Here, we follow a Euclidean based clustering as the segmentation algorithm.
We do not explore global features as they are highly point of view dependent.

\subsection{Descriptors}
This module takes each feature and computes a descriptor with the aim of being descriptive enough such that it is reproducible on different maps of the same location. 
The descriptor is based on the key-point, and its neighborhood, or on the subset of points that belong to a segment.
Here, we explore three descriptors:
\begin{itemize}
\item Fast Point Feature Histogram (\ac{FPFH})~\cite{rusu2009fast}.
\item Unique Signatures of Histograms for Local Surface Description (\ac{SHOT})~\cite{tombari2010unique}.
\item Ensemble of Shape Functions (\ac{ESF})~\cite{wohlkinger2011ensemble}.
\end{itemize}

\subsection{Description Matching}
The matching module is in charge of solving the data association problem between features from both maps by comparing their descriptors.
In our implementation we use the nearest neighbor search in the space of the corresponding descriptor.

\subsection{Transformation Estimation}
Once a set of 3D point pairs is declared, this module computes the transformation such that the 3D points from one map are moved to the location of their correspondences in the reference map.
In absence of outliers, the problem could be solved by minimizing a least square error function.
Unfortunately, the presence of outliers is unavoidable and this module must deal with them.
Here we explore two alternative methods.
The first one is a RANSAC-based approach which is already available in PCL.
The second one is based on the recent proposed \ac{FGR}~\cite{zhou2016fast}.
\ac{FGR} uses the scaled Geman-McClure estimator as robust cost function into the optimization objective to neutralize the possible outlier matches.

\subsection{Realizations}
\label{sec:realizations}
We explore different global registration alternatives by choosing different methods in each module.
The realizations are as shown in Table~\ref{tab:realizations}.
\begin{table}
\centering
\begin{tabular}{l c c c}
Realization & Feature & Descriptor & Trans. Estimation \\
\hline
FPFH & Key-point & \ac{FPFH} & RANSAC-based \\
FPFH FGR & Key-point & \ac{FPFH} & \ac{FGR} \\
FPFH seg & Segments & \ac{FPFH} & RANSAC-based \\
SHOT & Key-point & \ac{SHOT} & RANSAC-based \\
SHOT FGR & Key-point & \ac{SHOT} & \ac{FGR} \\
SHOT seg & Segments & \ac{SHOT} & RANSAC-based \\
ESF seg~\cite{dube2017icra} & Segments & \ac{ESF} & RANSAC-based \\
\end{tabular}
\caption{Global registration realization by different choices in the sub-modules.}
\label{tab:realizations}
\end{table}

As global registration strategies, the evaluation focuses on 11 different configurations.
Those that are shown in Table~\ref{tab:realizations} plus their combinations when removing the ground plane prior to key-point detection, denoted by \emph{gr} at the end.
Ground removal is done by RANSAC based plane fitting.
\subsection{Performance metrics}
\label{sec:performance_metrics}
For the evaluation metrics, we use transformation errors $\Delta \boldsymbol{T}$ on the alignment between the \ac{UGV} and \ac{UAV} maps that are represented as
\begin{equation}
\Delta \boldsymbol{T} = \begin{bmatrix}
\Delta \boldsymbol{R} & \Delta \boldsymbol{t} \\ \boldsymbol{0} & 1
\end{bmatrix}
\end{equation}
with rotation matrix $\Delta \boldsymbol{R}$ and translation vector $\Delta \boldsymbol{t} = (\Delta\mathbf{x}, \Delta\mathbf{y}, \Delta\mathbf{z})^{\mathbf{T}}$.
The translational error $e_t$ is computed as follows:
\begin{equation}
e_t = \norm{\Delta \boldsymbol{t}} = \sqrt{\Delta\mathbf{x}^2 + \Delta\mathbf{y}^2 + \Delta\mathbf{z}^2}
\end{equation}
The rotational error $e_r$ equates to:
\begin{equation}
e_r = \arccos{\frac{trace(\Delta \boldsymbol{R} - \boldsymbol{I})}{2}}
\end{equation}
It is important to note that the two map types are not perfectly aligned in all locations due to the multi-modal nature of the data and we can therefore only evaluate errors down to a positional resolution of $0.2m$ and angular resolution of $2^{\circ}$ respectively.
Furthermore, we register the data using ICP in the basic experiment, to give an indication on the achievable alignment, as ICP always converged to a good solution in our experiments given a good initial guess.
Motivated by the results of \citet{Hinzmann2016}, we consider registrations as successful when ICP is able to perform the final local alignment.
Therefore, the thresholds for translational and rotational errors are set to $e_{t} = 3m$ and $e_{r} = 5^{\circ}$ above the resulting ICP solution of the basic experiment to count successful registrations.

\section{Experiments}
We evaluate our approach on two challenging \ac{SaR} datasets recorded within the TRADR project which we make available with this publication\footnote{The datasets are available under \href{http://robotics.ethz.ch/~asl-datasets/}{http://robotics.ethz.ch/~asl-datasets/}}.
The evaluation focuses on the global registration of the multi-modal point-cloud data.
\subsection{Datasets}
\begin{figure}
\begin{subfigure}{0.95\columnwidth}
\begin{center}
\includegraphics[width = 0.95\columnwidth]{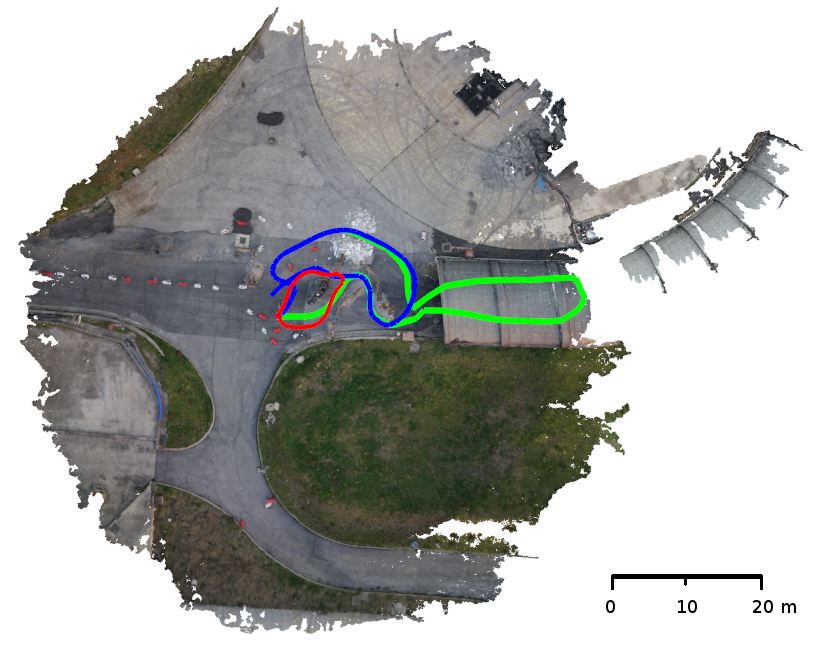}
\caption{Montelibretti outdoor dataset.}
\label{fig:montelibretti_paths}
\end{center}
\end{subfigure}

\medskip
\begin{subfigure}{0.95\columnwidth}
\begin{center}
\includegraphics[width = 0.95\columnwidth]{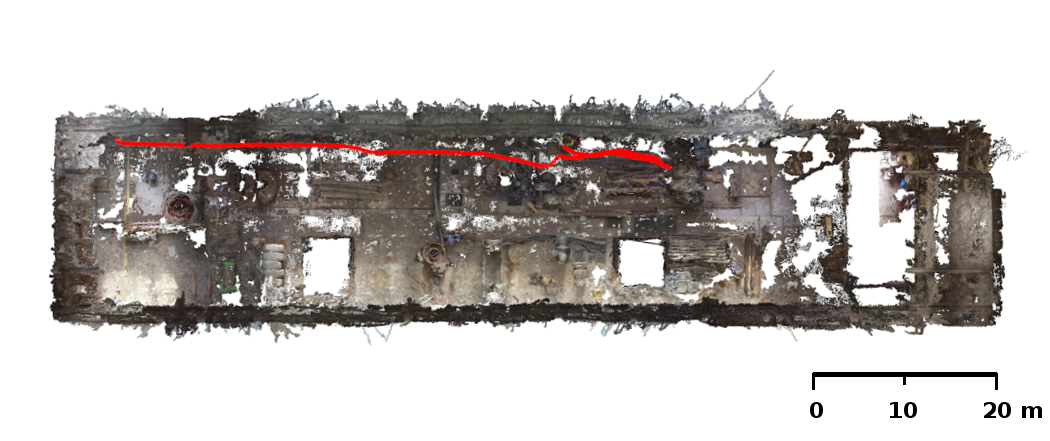}
\caption{Dortmund indoor dataset.}
\label{fig:knepper_paths}
\end{center}
\end{subfigure}
\caption{Top-down views of two \ac{SaR} datasets considered in our experiments. The robot trajectories are indicated in green, red, and blue and overlaid on the colored \ac{UAV} point-clouds.}
\end{figure}

The first of these datasets was generated in an outdoor firemen training location in Montelibretti, Italy.
The scenario simulates a car accident around a tunnel.
It consists of six \ac{UGV} runs in partly overlapping locations of the disaster area and one large \ac{UAV}-generated map covering the whole site of approximately $80m \times 80m$ that was scaled using GPS information.
The travelled trajectories of the robots are $2 \times 30m$, $2 \times 60m$, and $2 \times 110m$ of consecutive missions following the same paths twice.
For the evaluation we use one \ac{UGV} run of each size.
Here, the \ac{UGV} mapping data is fully covered in the \ac{UAV} map, except for an indoor exploration of the tunnel which was not accessible to the \ac{UAV}.
The scenario and the trajectories are depicted in Fig.~\ref{fig:montelibretti_paths}.

The second dataset was recorded at a decommissioned power plant in Dortmund, Germany, consisting of several \ac{UGV} runs and several large \ac{UAV}-generated maps covering different parts of the power plant, including the entire machine hall which was also visited by one \ac{UGV} and has a size of $100m \times 20m$.
The \ac{UGV} run is fully covered in the \ac{UAV} map and has a travelled distance of approximately $80m$, as illustrated in Fig.~\ref{fig:knepper_paths}.
Since no GPS signals are available in the interior of the building, the \ac{UAV} maps were scaled using the buildings' windows as reference to the outside maps and scaled accordingly.
\subsection{Experimental setup}
We evaluate several global registration strategies on increasingly challenging experimental set-ups.
All set-ups consider the iteratively growing \ac{UGV} map produced by laserSLAM as local map.
For the basic set-up, the global map is a cropped version of the \ac{UAV} map, that approximately covers the space of the local map at any iteration.
The more challenging intermediate set-up uses the cropped \ac{UAV} map as seen at the last iteration of the \ac{UGV} mapping as global map.
Finally, the complex set-up considers the full \ac{UAV} map for global localization.

As global registration strategies, the evaluation focuses on the 11 configurations presented in Sec.~\ref{sec:realizations}.
\subsection{Registration performance}
This section evaluates the global registration performance of the different algorithms considered, using the metrics presented in Sec.~\ref{sec:performance_metrics}.
\subsubsection{Parametrization}
We choose the parametrization of the \ac{FPFH} and \ac{SHOT} descriptors to yield good performance across all data used, i.e., a histogram and search radius of $2.0m$ for \ac{FPFH} and \ac{SHOT} respectively.
Our parameter choice is further motivated by extensive evaluation and shows plateauing performance in a large region around the chosen size, indicating robust performance.
The matcher is based on performing fast nearest neighbor search in a FLANN tree \cite{muja2009fast}, while the geometric verification is based on RANSAC and clustering.
%
Furthermore, \ac{FGR} is parametrized for the best possible performance we could find.
\subsection{Results}
\begin{figure}
\centering
\includegraphics[width = 0.95\columnwidth]{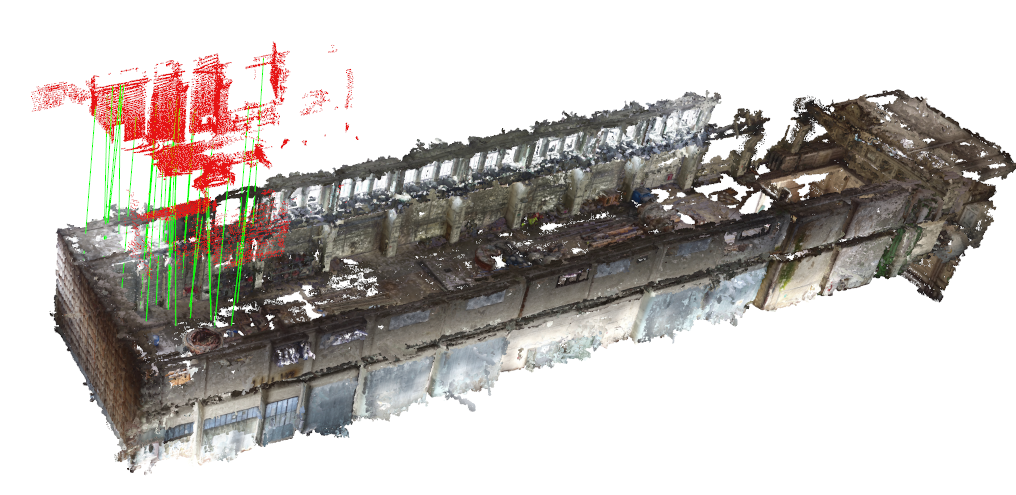}
\caption{Resulting global localization between \emph{3D} \ac{UGV} sub-map (red point-cloud) and global \emph{3D} \ac{UAV} map (coloured point-cloud). Green lines indicate the resulting descriptor matches associated to points in the two point-clouds. The data stems from the Dortmund indoor dataset and drawn from the complex experimental set-up.}
\label{fig:knepper}
\end{figure}

\begin{figure*}
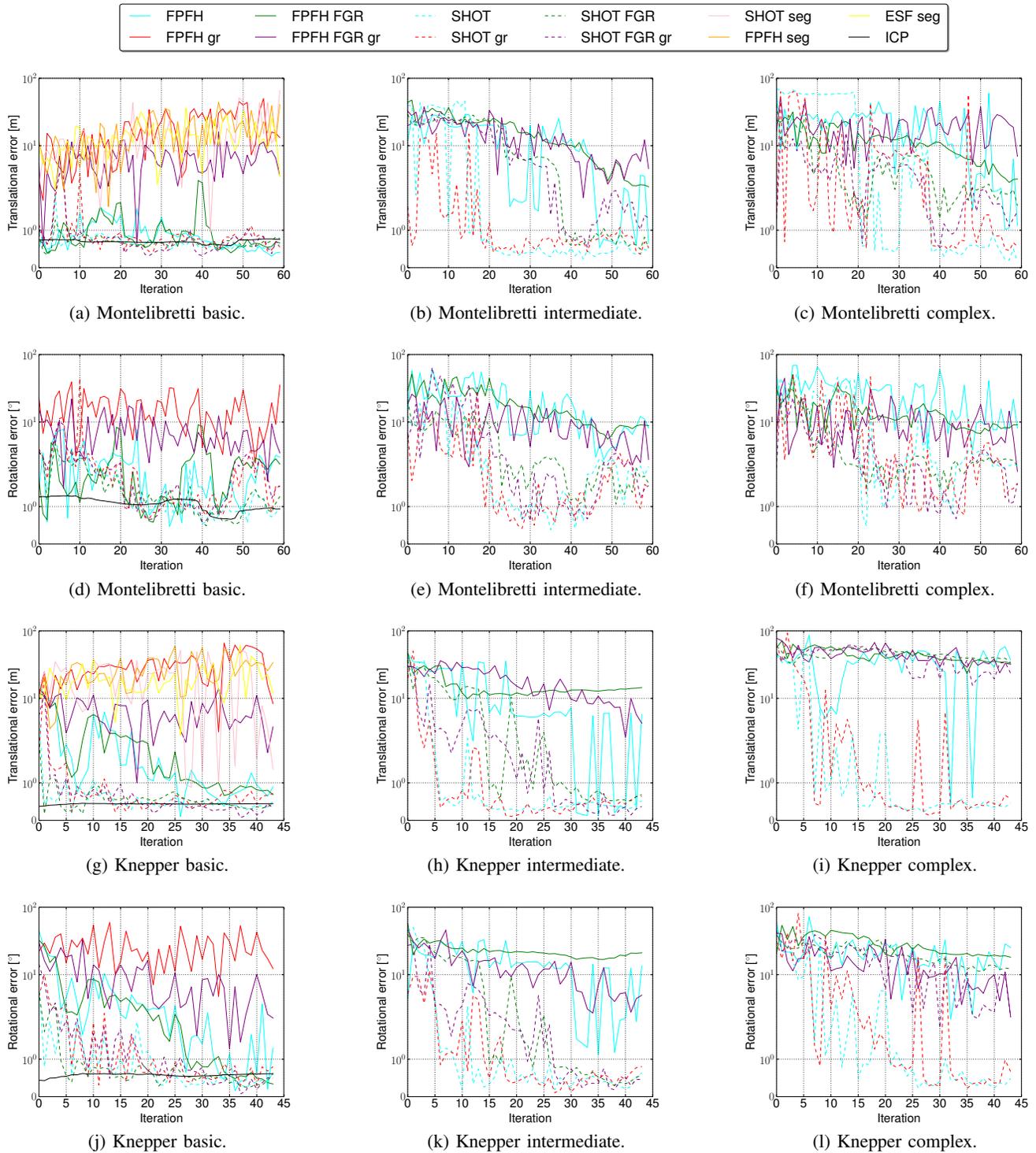

\begin{subfigure}{0.3\textwidth}
\resizebox{\textwidth}{!}{\input{figures/trans_error_montelibretti_basic.pgf}}
\caption{Montelibretti basic.}
\label{fig:trans_montelibretti_basic}
\end{subfigure}\hspace*{\fill}
\begin{subfigure}{0.3\textwidth}
\begin{center}
\resizebox{\textwidth}{!}{\input{figures/trans_error_montelibretti_intermediate.pgf}}
\caption{Montelibretti intermediate.}
\label{fig:trans_montelibretti_intermediate}
\end{center}
\end{subfigure}\hspace*{\fill}
\begin{subfigure}{0.3\textwidth}
\begin{center}
\resizebox{\textwidth}{!}{\input{figures/trans_error_montelibretti_complex.pgf}}
\caption{Montelibretti complex.}
\label{fig:trans_montelibretti_complex}
\end{center}
\end{subfigure}
\smallbreak
\begin{subfigure}{0.3\textwidth}
\resizebox{\textwidth}{!}{\input{figures/rot_error_montelibretti_basic.pgf}}
\caption{Montelibretti basic.}
\label{fig:rot_montelibretti_basic}
\end{subfigure}\hspace*{\fill}
\begin{subfigure}{0.3\textwidth}
\begin{center}
\resizebox{\textwidth}{!}{\input{figures/rot_error_montelibretti_intermediate.pgf}}
\caption{Montelibretti intermediate.}
\label{fig:rot_montelibretti_intermediate}
\end{center}
\end{subfigure}\hspace*{\fill}
\begin{subfigure}{0.3\textwidth}
\begin{center}
\resizebox{\textwidth}{!}{\input{figures/rot_error_montelibretti_complex.pgf}}
\caption{Montelibretti complex.}
\label{fig:rot_montelibretti_complex}
\end{center}
\end{subfigure}
\smallbreak
\begin{subfigure}{0.3\textwidth}
\resizebox{\textwidth}{!}{\input{figures/trans_error_knepper_basic.pgf}}
\caption{Knepper basic.}
\label{fig:trans_knepper_basic}
\end{subfigure}\hspace*{\fill}
\begin{subfigure}{0.3\textwidth}
\begin{center}
\resizebox{\textwidth}{!}{\input{figures/trans_error_knepper_intermediate.pgf}}
\caption{Knepper intermediate.}
\label{fig:trans_knepper_intermediate}
\end{center}
\end{subfigure}\hspace*{\fill}
\begin{subfigure}{0.3\textwidth}
\begin{center}
\resizebox{\textwidth}{!}{\input{figures/trans_error_knepper_complex.pgf}}
\caption{Knepper complex.}
\label{fig:trans_knepper_complex}
\end{center}
\end{subfigure}
\smallbreak
\begin{subfigure}{0.3\textwidth}
\resizebox{\textwidth}{!}{\input{figures/rot_error_knepper_basic.pgf}}
\caption{Knepper basic.}
\label{fig:rot_knepper_basic}
\end{subfigure}\hspace*{\fill}
\begin{subfigure}{0.3\textwidth}
\begin{center}
\resizebox{\textwidth}{!}{\input{figures/rot_error_knepper_intermediate.pgf}}
\caption{Knepper intermediate.}
\label{fig:rot_knepper_intermediate}
\end{center}
\end{subfigure}\hspace*{\fill}
\begin{subfigure}{0.3\textwidth}
\begin{center}
\resizebox{\textwidth}{!}{\input{figures/rot_error_knepper_complex.pgf}}
\caption{Knepper complex.}
\label{fig:rot_knepper_complex}
\end{center}
\end{subfigure}
\caption{Translational and rotational error of plots for the Montelibretti (outdoor) and Dortmund (indoor) experiments on global registration. All errors are plotted over the number of iterations, i.e., the growing size of the \ac{UGV} map. Here, we indicate experimental configurations as follows: \emph{SHOT}, \emph{FPFH} and \emph{ESF} denote the used descriptors, an additional \emph{FGR} denotes if we used the fast global optimization instead of the RANSAC-based outlier filtering, and we add \emph{gr} for cases in which also ground removal was performed before descriptor extraction. The \emph{ICP} solution is illustrated for the basic experiments. Plots~\subref{fig:trans_montelibretti_basic}-~\subref{fig:trans_montelibretti_complex} illustrate the translational errors for the Montelibretti experiment for the basic, indermediate, and complex set-up, while plots~\subref{fig:trans_knepper_basic}-~\subref{fig:trans_knepper_complex} show the translational errors for the respective experiments on the Dortmund data. Figures~\subref{fig:rot_montelibretti_basic}-~\subref{fig:rot_montelibretti_complex} illustrate the rotational errors for the Montelibretti experiment and Figures~\subref{fig:rot_knepper_basic}-~\subref{fig:rot_knepper_complex} the respective rotational errors on the Dortmund data.} 
\label{fig:error_plots}
\end{figure*}

Fig.~\ref{fig:montelibretti} and Fig.~\ref{fig:knepper} illustrate qualitative global registration between \ac{UGV} sub-maps and global \ac{UAV} maps on the tested datasets.
In Fig.~\ref{fig:error_plots} and Table~\ref{tab:performance} the quantitative performance of the evaluated approaches is depicted as averaged over multiple runs with different initializations.
While Fig.~\ref{fig:error_plots} shows translational and rotational error of the individual approaches over all datasets, Table~\ref{tab:performance} reports the minimal amount of cumulated \ac{UGV} scans, i.e., the minimal travelled distance for reliable global registration.
Here, we define reliable global registration, if from the associated \ac{UGV} sub-map, the errors do not exceed the error thresholds $e_t$ and $e_r$ for $90\%$ of the cases.
Note, that we indicate combinations that failed to produce result within this margin as \emph{N/A}.

In this experimental set-up the descriptor matching approach as described in Section~\ref{sec:global_registration} performs best throughout all experiments.
\ac{FPFH} yields satisfying performance in the basic experiments.
However, its performance drastically degrades in the more complex cases.
%

\ac{SHOT} on the other hand shows reliable performance throughout all experiments, with the required overlap increasing with the complexity of our test-cases.
Here, the ground removal does not provide a significant performance boost, especially since the ground plane extraction is unreliable on the large \ac{UAV} map as we do not have incremental pose-updates from the robot on segments of the map.
However, ground removal does not degrade the performance as it does for \ac{FPFH}, expressing robust performance of \ac{SHOT} over varying conditions.

While the RANSAC-based geometric verification can reject a large amount of mismatched descriptors and does not rely on the point of initialization, \ac{FGR} is less robust to poor initialization as done for the intermediate and complex experiments.
For the reduced search space in the basic and intermediate experiments, \ac{FGR} is able to achieve reasonable registration performance and therefore shows high potential to be used for such reduced search problems, when carefully modelling its robust cost function.

While the segmentation approach shows very good performance for reproducible segmentations, e.g., single-modality localization~\cite{dube2017icra}, we found that the considered parametrizations of Euclidean segmentation in combination with the considered descriptors did not generalize well between the modalities and could not deliver interesting results in the experiments.
For the sake of clarity of the plots, we therefore only show their performance in the basic experiments in Fig.~\ref{fig:error_plots}.
Since the remainder of the matching algorithm is identical to the well performing descriptor matching, we believe that given a reliable segmentation, the approach has the potential to yield very good performance for the global registration case.
However, a purely geometric ground removal and segmentation on the full \ac{UAV} point-cloud that is comparable to the segmentation on the \ac{UGV} map is a hard problem, especially for cluttered \ac{SaR} scenarios.
%

%
\begin{table}
\centering
\begin{tabular}{l*{6}{c}r}
Configuration & $M_b$ & $M_i$ & $M_c$ & $D_b$ & $D_i$ & $D_c$ \\
\hline
FPFH & 1 & 45 & 60 & 2 & N/A & N/A  \\
FPFH gr & N/A & N/A & N/A & N/A & N/A &  N/A  \\
FPFH FGR & 1 & 50 & 54 & 3 & N/A &  N/A  \\
FPFH FGR gr & 60 & 60 & 43 & 3 & N/A &  N/A  \\
SHOT & 1 & 10 & 36 & 1 & 1 & 3  \\
SHOT gr & 1 & 1 & 36 & 1 &  1 &  7  \\
SHOT FGR & 1 & 35 & 35 & 1 & 12 &  N/A  \\
SHOT FGR gr & 1 & 29 & 28 & 1 & 12 &  N/A  \\
FPFH seg & N/A & N/A & N/A & N/A & N/A &  N/A  \\
SHOT seg & N/A & N/A & N/A & N/A & N/A &  N/A  \\
ESF seg & N/A & N/A & N/A & N/A & N/A &  N/A  \\
\end{tabular}
\caption{Minimal number of LiDAR scans for successful global registration experiments. Here, $M_b$, $M_i$, $M_c$, denote the basic, intermediate, and complex experiment on the Montelibretti dataset, while $D_b$, $D_i$, and $D_c$ denote the different experimental set-ups on the Dortmund data. The \ac{UGV} travels on average $0.8m$ between two scans.}
\label{tab:performance}
\end{table}
\subsubsection{Timings}
Additionally to the evaluation of residuals, computation times are an important factor in the choice of algorithms in robotics.
Table~\ref{tab:times} lists the computational times of the four main components of the global registration algorithms when executed on an Intel i7-4600U CPU @ 2.10GHz.
The computation times are reported per \ac{UGV} LiDAR scan which has an acquisition time of $3s$ on the considered platform.

Although our focus was not on maximizing computational efficiency, all approaches can be performed in this time window and therewith in real-time.
We are confident that they can be further improved to also yield faster processing times.
With key-point detection times increasing with the amount of points in the point-cloud, the segmentation approach is the fastest in the first step.
Descriptor extraction, is fastest for \ac{SHOT} descriptors, also scaling with the amount of points.
However, the largest contribution to the computational time has the descriptor matching which is longest for \ac{SHOT} and low for \ac{FPFH}.
For the segmentation, we report the timings for the high-dimensional \ac{ESF} features and achieve low timings, due to the compact representation of segments.
Finally, RANSAC-based geometric consistency is slower than the optimization-based \ac{FGR}.
\setlength{\belowcaptionskip}{-10pt}
\begin{table*}
\centering
\begin{tabular}{c|ccccccccc|}
Module & \thead{\emph{FPFH}\\ } & \thead{\emph{FPFH} \\ \emph{gr}} & \thead{\emph{FPFH} \\ \emph{FGR}} & \thead{\emph{FPFH} \\ \emph{FGR gr}} & \thead{\emph{SHOT}\\ } & \thead{\emph{SHOT}\\ \emph{gr}} & \thead{\emph{SHOT}\\ \emph{FGR}} & \thead{\emph{SHOT}\\ \emph{FGR gr}} & \thead{\emph{Seg}\\ }\\
\hline
\thead{Key-point / \\ Segmentation} & \thead{$35.97$\\$\pm 13.17$} & \thead{$22.92$\\$\pm 13.30$} & \thead{$39.01$\\$\pm 16.04$} & \thead{$24.57$\\$\pm 14.90$} & \thead{$33.1$\\$\pm 12.16$} & \thead{$18.97$\\$\pm 11.07$} & \thead{$35.76$\\$\pm 13.30$} & \thead{$23.06$\\$\pm 14.87$} & \thead{$9.32$\\$\pm 5$}\\
Description & \thead{$228.87$\\$\pm 61.99$} & \thead{$122.59$\\$\pm 39.27$} & \thead{$242.52$\\$\pm 64.34$} & \thead{$136.63$\\$\pm 43.91$} & \thead{$17.21$\\$\pm 8.82$} & \thead{$12.30$\\$\pm 8.60$} & \thead{$17.35$\\$\pm 8.51$} & \thead{$13.52$\\$\pm 9.52$} & \thead{$105.91$\\$\pm 50.72$} \\
Matching & \thead{$293.32$ \\ $\pm 8.32$} & \thead{$86.04$\\$\pm 27.41$} & \thead{$291.76$ \\ $\pm 6.49$} & \thead{$88.33$\\$\pm 25.25$} & \thead{$2897.94$\\$\pm 44.01$} & \thead{$1992.17$\\$\pm 40.36$} & \thead{$2937.60$\\$\pm 65.13$} & \thead{$2099.55$\\$\pm 42.31$} & \thead{$429.11$\\$\pm 51.15$} \\
\thead{Geometric \\ consistency / \\ Optimization} & \thead{$15.91$\\$\pm 11.54$} & \thead{$22.72$\\$\pm 17.28$} & \thead{$2.38$\\$\pm 1.09$} & \thead{$2.05$\\$\pm 1.05$} & \thead{$7.55$\\$\pm 6.11$} & \thead{$8.24$\\$\pm 6.79$} & \thead{$2.23$\\$\pm 0.99$} & \thead{$1.82$\\$\pm 0.94$} & \thead{$0.80$\\$\pm 1.60$} \\
\hline
Total & $574.07$ & $254.27$ & $575.67$ & $251.58$ & $2926.01$ & $2031.68$ & $2992.94$ & $2137.95$ & $545.14$  \\
\end{tabular}
\caption{Mean computation times and standard deviations of the individual approaches in the complex Montelibretti experiment per LiDAR scan in $ms$ as computed on a single core of an Intel i7-4600U CPU @ 2.10GHz.}
\label{tab:times}
\end{table*}
\subsection{Discussion}
The global registration of \emph{3D} \ac{UAV} and \ac{UGV} point-cloud data is a difficult problem.
Based on our evaluation, the most general solution that we devise is a key-point descriptor matching algorithm using \ac{SHOT} descriptors.
In our evaluation, \ac{FPFH} descriptors performed well, with large overlap between the maps, but failed for the more complex experiments, and showed to be sensitive to ground removal.

The segmentation showed to not deliver satisfying results, as it requires repeatable ground removal and segmentation, which could not be achieved in the considered configurations and scenarios.
Key-point detection on the other hand performed well.

\ac{SHOT} descriptor matching is computationally more expensive than \ac{FPFH} due to high descriptor dimensionality, but showed best performance throughout.
The processing time can be speeded up by removing the ground in environments that allow for reliable ground removal.

Finally \ac{FGR} can yield additional speed up of the transformation estimation.
Yet, when using \ac{FGR}, the cost function must be carefully considered as the technique is prone to converge to local minima.
While the RANSAC-based transformation estimation takes generally longer than \ac{FGR} the robustness to local minima is greatly increased.
Also, the additional computational time for RANSAC was negligible in our experiments.
\section{Conclusion}
This paper presented global registration algorithms for \ac{UGV} and \ac{UAV} point-clouds generated from heterogeneous sensors, i.e., LiDAR sensors for \ac{UGV}s and cameras for \ac{UAV}s, and drastically different view-points.
The registration algorithm is based on geometrical descriptor matching.
The approach was integrated with a full \ac{SaR} robotic mapping system, bridging the gap between effective exploitation of \ac{UAV} mapping data on \ac{UGV}s.
We evaluated several different \emph{3D} descriptor-based registration techniques and identify the best performing approach for the problem of global point-cloud registration from heterogeneous sensors in \ac{SaR} scenarios.

Future avenues of research could include point-cloud registration by using further informative cues than the geometrical information alone for data registration between the sensor modalities.
This could benefit runtime and compactness of point-cloud description of the proposed algorithm.
\section{Acknowledgement}
This work was supported by European Union's Seventh Framework Programme for research, technological development and demonstration under the TRADR project No. FP7-ICT-609763.

\bibliographystyle{IEEEtranN}

\bibliography{eth}

\begin{thebibliography}{24}
\providecommand{\natexlab}[1]{#1}
\providecommand{\url}[1]{#1}
\csname url@samestyle\endcsname
\providecommand{\newblock}{\relax}
\providecommand{\bibinfo}[2]{#2}
\providecommand{\BIBentrySTDinterwordspacing}{\spaceskip=0pt\relax}
\providecommand{\BIBentryALTinterwordstretchfactor}{4}
\providecommand{\BIBentryALTinterwordspacing}{\spaceskip=\fontdimen2\font plus
\BIBentryALTinterwordstretchfactor\fontdimen3\font minus
  \fontdimen4\font\relax}
\providecommand{\BIBforeignlanguage}[2]{{%
\expandafter\ifx\csname l@#1\endcsname\relax
\typeout{** WARNING: IEEEtranN.bst: No hyphenation pattern has been}%
\typeout{** loaded for the language `#1'. Using the pattern for}%
\typeout{** the default language instead.}%
\else
\language=\csname l@#1\endcsname
\fi
#2}}
\providecommand{\BIBdecl}{\relax}
\BIBdecl

\bibitem[Saeedi et~al.(2016)Saeedi, Trentini, Seto, and Li]{saeedi2016multiple}
S.~Saeedi, M.~Trentini, M.~Seto, and H.~Li, ``Multiple-robot simultaneous
  localization and mapping: A review,'' \emph{JFR}, pp. 3--46, 2016.

\bibitem[Michael et~al.(2012)Michael, Shen, Mohta, Mulgaonkar, Kumar, Nagatani,
  Okada, Kiribayashi, Otake, Yoshida, et~al.]{michael2012collaborative}
N.~Michael, S.~Shen, K.~Mohta, Y.~Mulgaonkar, V.~Kumar, K.~Nagatani, Y.~Okada,
  S.~Kiribayashi, K.~Otake, K.~Yoshida \emph{et~al.}, ``Collaborative mapping
  of an earthquake-damaged building via ground and aerial robots,'' \emph{JFR},
  pp. 832--841, 2012.

\bibitem[Kim et~al.(2010)Kim, Kaess, Fletcher, Leonard, Bachrach, Roy, and
  Teller]{Kim10icra}
B.~Kim, M.~Kaess, L.~Fletcher, J.~J. Leonard, A.~Bachrach, N.~Roy, and
  S.~Teller, ``{Multiple Relative Pose Graphs for Robust Cooperative
  Mapping},'' in \emph{ICRA}, 2010, pp. 3185--3192.

\bibitem[Zeng et~al.(2017)Zeng, Song, Nie{\ss}ner, Fisher, Xiao, and
  Funkhouser]{zeng2017cvpr}
A.~Zeng, S.~Song, M.~Nie{\ss}ner, M.~Fisher, J.~Xiao, and T.~Funkhouser,
  ``3dmatch: Learning local geometric descriptors from rgb-d reconstructions,''
  in \emph{CVPR}, 2017.

\bibitem[Dub\'e et~al.(2017{\natexlab{a}})Dub\'e, Gawel, Nieto, Siegwart, and
  Cadena]{Dube2017iros}
R.~Dub\'e, A.~Gawel, J.~Nieto, R.~Siegwart, and C.~Cadena, ``Online multi-robot
  slam with 3d lidars: A full system,'' in \emph{IROS}, 2017.

\bibitem[Fuhrmann et~al.(2014)Fuhrmann, Langguth, and Goesele]{fuhrmann2014mve}
S.~Fuhrmann, F.~Langguth, and M.~Goesele, ``Mve-a multi-view reconstruction
  environment.'' in \emph{GCH}, 2014, pp. 11--18.

\bibitem[Birk and Carpin(2006)]{birk2006merging}
A.~Birk and S.~Carpin, ``Merging occupancy grid maps from multiple robots,''
  \emph{Proceedings of the IEEE}, pp. 1384--1397, 2006.

\bibitem[Blanco et~al.(2013)Blanco, Gonz{\'a}lez-Jim{\'e}nez, and
  Fern{\'a}ndez-Madrigal]{blanco2013robust}
J.-L. Blanco, J.~Gonz{\'a}lez-Jim{\'e}nez, and J.-A. Fern{\'a}ndez-Madrigal,
  ``A robust, multi-hypothesis approach to matching occupancy grid maps,''
  \emph{Robotica}, pp. 687--701, 2013.

\bibitem[Saeedi et~al.(2011)Saeedi, Paull, Trentini, and
  Li]{saeedi2011multiple}
S.~Saeedi, L.~Paull, M.~Trentini, and H.~Li, ``Multiple robot simultaneous
  localization and mapping,'' in \emph{IROS}, 2011, pp. 853--858.

\bibitem[Cadena et~al.(2016)Cadena, Carlone, Carrillo, Latif, Scaramuzza,
  Neira, Reid, and Leonard]{Cadena16tro-SLAMfuture}
C.~Cadena, L.~Carlone, H.~Carrillo, Y.~Latif, D.~Scaramuzza, J.~Neira, I.~Reid,
  and J.~Leonard, ``Past, present, and future of simultaneous localization and
  mapping: Towards the robust-perception age,'' \emph{IEEE Transactions on
  Robotics}, pp. 1309--1332, 2016.

\bibitem[Forster et~al.(2013)Forster, Pizzoli, and Scaramuzza]{forster2013air}
C.~Forster, M.~Pizzoli, and D.~Scaramuzza, ``Air-ground localization and map
  augmentation using monocular dense reconstruction,'' in \emph{IROS}, 2013,
  pp. 3971--3978.

\bibitem[Hinzmann et~al.(2016)Hinzmann, Stastny, Conte, Doherty, Rudol, Wzorek,
  Galceran, Siegwart, and Gilitschenski]{Hinzmann2016}
T.~Hinzmann, T.~Stastny, G.~Conte, P.~Doherty, P.~Rudol, M.~Wzorek,
  E.~Galceran, R.~Siegwart, and I.~Gilitschenski, \emph{Collaborative 3D
  Reconstruction Using Heterogeneous UAVs: System and Experiments}.\hskip 1em
  plus 0.5em minus 0.4em\relax Cham: Springer International Publishing, 2016,
  pp. 43--56.

\bibitem[Wolcott and Eustice(2014)]{wolcott2014visual}
R.~W. Wolcott and R.~M. Eustice, ``Visual localization within lidar maps for
  automated urban driving,'' in \emph{IROS}, 2014, pp. 176--183.

\bibitem[Gawel et~al.(2016)Gawel, Cieslewski, Dub{\'e}, Bosse, Siegwart, and
  Nieto]{gawel2016structure}
A.~Gawel, T.~Cieslewski, R.~Dub{\'e}, M.~Bosse, R.~Siegwart, and J.~Nieto,
  ``Structure-based vision-laser matching,'' in \emph{IROS}, 2016, pp.
  182--188.

\bibitem[Dub\'e et~al.(2017{\natexlab{b}})Dub\'e, Dugas, Stumm, Nieto,
  Siegwart, and Cadena]{dube2017icra}
R.~Dub\'e, D.~Dugas, E.~Stumm, J.~Nieto, R.~Siegwart, and C.~Cadena,
  ``\textit{SegMatch}: Segment based place recognition in 3d point clouds,'' in
  \emph{ICRA}, 2017.

\bibitem[Zhou et~al.(2016)Zhou, Park, and Koltun]{zhou2016fast}
Q.-Y. Zhou, J.~Park, and V.~Koltun, ``Fast global registration,'' in
  \emph{ECCV}, 2016, pp. 766--782.

\bibitem[Hornung et~al.(2013)Hornung, Wurm, Bennewitz, Stachniss, and
  Burgard]{hornung2013octomap}
A.~Hornung, K.~M. Wurm, M.~Bennewitz, C.~Stachniss, and W.~Burgard, ``Octomap:
  An efficient probabilistic 3d mapping framework based on octrees,''
  \emph{AuRo}, pp. 189--206, 2013.

\bibitem[Dub{\'e} et~al.(2016)Dub{\'e}, Gawel, Cadena, Siegwart, Freda, and
  Gianni]{dube20163d}
R.~Dub{\'e}, A.~Gawel, C.~Cadena, R.~Siegwart, L.~Freda, and M.~Gianni, ``3d
  localization, mapping and path planning for search and rescue operations,''
  in \emph{SSRR}, 2016, pp. 272--273.

\bibitem[Kaess et~al.(2012)Kaess, Johannsson, Roberts, Ila, Leonard, and
  Dellaert]{kaess2012isam2}
M.~Kaess, H.~Johannsson, R.~Roberts, V.~Ila, J.~J. Leonard, and F.~Dellaert,
  ``isam2: Incremental smoothing and mapping using the bayes tree,''
  \emph{IJRR}, pp. 216--235, 2012.

\bibitem[Zhong(2009)]{zhong2009intrinsic}
Y.~Zhong, ``Intrinsic shape signatures: A shape descriptor for 3d object
  recognition,'' in \emph{ICCV}, 2009, pp. 689--696.

\bibitem[Rusu et~al.(2009)Rusu, Blodow, and Beetz]{rusu2009fast}
R.~B. Rusu, N.~Blodow, and M.~Beetz, ``Fast point feature histograms (fpfh) for
  3d registration,'' in \emph{ICRA}, 2009, pp. 3212--3217.

\bibitem[Tombari et~al.(2010)Tombari, Salti, and Di~Stefano]{tombari2010unique}
F.~Tombari, S.~Salti, and L.~Di~Stefano, ``Unique signatures of histograms for
  local surface description,'' in \emph{ECCV}, 2010, pp. 356--369.

\bibitem[Wohlkinger and Vincze(2011)]{wohlkinger2011ensemble}
W.~Wohlkinger and M.~Vincze, ``Ensemble of shape functions for 3d object
  classification,'' in \emph{ROBIO}, 2011, pp. 2987--2992.

\bibitem[Muja and Lowe(2009)]{muja2009fast}
M.~Muja and D.~G. Lowe, ``Fast approximate nearest neighbors with automatic
  algorithm configuration.'' \emph{VISAPP}, pp. 331--340, 2009.

\end{thebibliography}

\begin{acronym}
\acro{ICP}{Iterative Closest Point}
\acro{UAV}{Unmanned Aerial Vehicle}
\acro{SLAM}{Simultaneous Localization and Mapping}
\acro{UGV}{Unmanned Ground Vehicle}
\acro{TRADR}{``Long-Term Human-Robot Teaming for Robots Assisted Disaster Response''}
\acro{SaR}{Search and Rescue}
\acro{IMU}{Inertial Measurement Unit}
\acro{SfM}{Structure from Motion}
\acro{FPFH}{Fast Point Feature Histogram}
\acro{SHOT}{Unique Signatures of Histograms for Local Surface Description}
\acro{ESF}{Ensemble of Shape Functions}
\acro{FGR}{Fast Global Registration}
\acro{MVE}{Multi-View Reconstruction Environment}
\acro{ISS}{Intrinsic Shape Signatures}
\end{acronym}

\end{document}